\pdfoutput=1

\documentclass[11pt]{article}

\usepackage{ACL2023}

\usepackage{times}
\usepackage{latexsym}
\usepackage{amsfonts}
\usepackage{amssymb}
\usepackage{amsmath}
\usepackage{multirow}
\usepackage{bm}
\usepackage{graphicx}
\usepackage{subfig}
\usepackage{booktabs}
\usepackage{enumitem}
\usepackage{float}

\usepackage[T1]{fontenc}

\usepackage[utf8]{inputenc}

\usepackage{microtype}

\usepackage{inconsolata}

\title{Infusing Hierarchical Guidance into Prompt Tuning: A Parameter-Efficient Framework for Multi-level Implicit Discourse Relation Recognition}

\author{Haodong Zhao\textsuperscript{1,2}, Ruifang He\textsuperscript{1,2}\thanks{\ \ The Corresponding author.}, Mengnan Xiao\textsuperscript{1,2} \and Jing Xu\textsuperscript{1,2} \\
        \textsuperscript{1}College of Intelligence and Computing, Tianjin University, Tianjin, China \\ \textsuperscript{2}Tianjin Key Laboratory of Cognitive Computing and Application, Tianjin, China\\ \texttt{\{haodongzhao,rfhe,mnxiao,jingxu\}@tju.edu.cn}}

\begin{document}
\maketitle
\begin{abstract}
Multi-level implicit discourse relation recognition (MIDRR) aims at identifying hierarchical discourse relations among arguments. Previous methods achieve the promotion through fine-tuning PLMs. However, due to the data scarcity and the task gap, the pre-trained feature space cannot be accurately tuned to the task-specific space, which even aggravates the collapse of the vanilla space. Besides, the comprehension of hierarchical semantics for MIDRR makes the conversion much harder. In this paper, we propose a prompt-based Parameter-Efficient Multi-level IDRR (PEMI) framework to solve the above problems. First, we leverage parameter-efficient prompt tuning to drive the inputted arguments to match the pre-trained space and realize the approximation with few parameters. Furthermore, we propose a hierarchical label refining (HLR) method for the prompt verbalizer to deeply integrate hierarchical guidance into the prompt tuning. Finally, our model achieves comparable results on PDTB 2.0 and 3.0 using about 0.1\% trainable parameters compared with baselines and the visualization demonstrates the effectiveness of our HLR method.
\end{abstract}

\section{Introduction}
\begin{figure}
    \centering
    \includegraphics[width=0.92\linewidth]{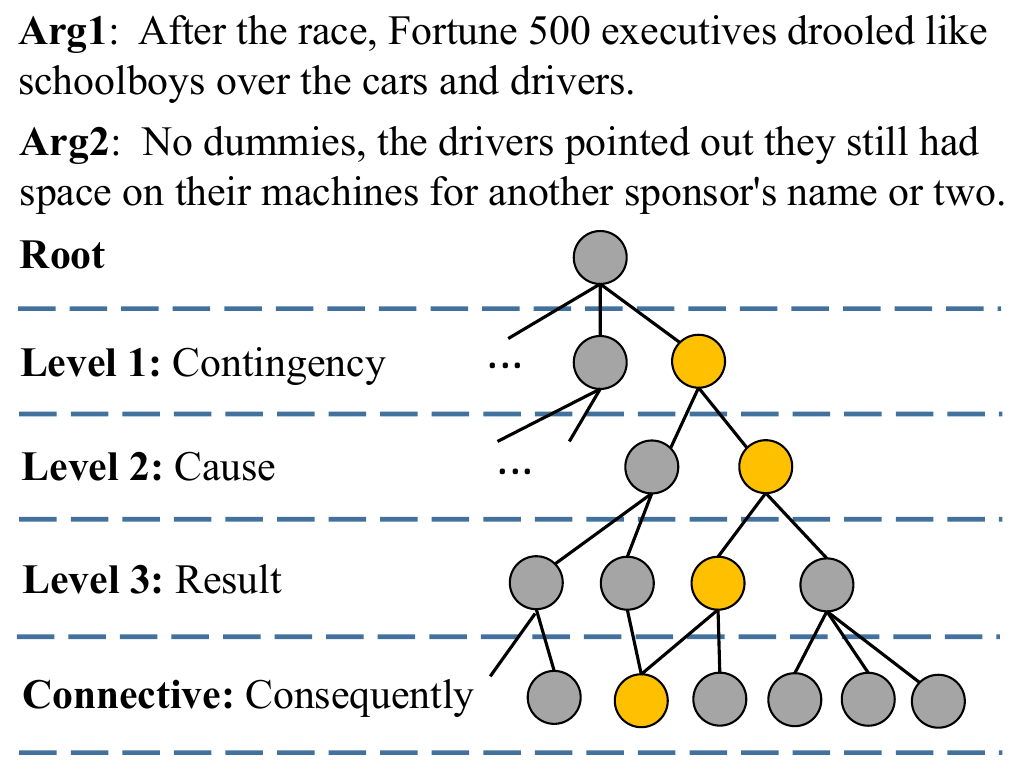}
    \caption{An Instance for multi-level IDRR.}
    \label{fig:figure1}
\end{figure}

Implicit discourse relation recognition (IDRR) \cite{Pitler2009} is one of the most vital sub-tasks in discourse analysis, which proposes to discover the discourse relation between two discourse arguments without the guidance of explicit connectives. Due to the lack of connectives, the model can only recognize the relations through semantic clues and entity anaphora between arguments, which makes IDRR a challenging task. Through a deeper understanding of this task, it is beneficial to a series of downstream tasks such as text summarization \cite{Li2020a}, dialogue summarization \cite{Feng2021} and event relation extraction \cite{Tang2021}.
Meanwhile, the discourse relation is annotated as multi-level labels. As shown in Figure \ref{fig:figure1}, the top-level label of this argument pair is \textit{Comparison}, while the sub-label \textit{Contrast} is the fine-grained semantic expression of \textit{Comparison}. Beyond that, when annotating the implicit relation, the annotators simulate adding a connective \textit{Consequently}. We regard these connectives as the bottom level of discourse relations. 

Since pre-trained language models (PLMs) are widely applied, IDRR has also achieved considerable improvement. However, previous work \cite{Xu2018, Shi2018, Dou2021} has mentioned the data scarcity of the IDRR, in which data is insufficient to support deep neural networks to depict the high-dimensional task-specific feature space accurately. Simultaneously, the hierarchical division of discourse relations is complex, and the extraction of hierarchical semantics relies on a large scale of data to sustain.

Previous studies \cite{Xu2018, Dai2019, Kishimoto2020, Guo2020, Shi2021} alleviate this problem by data augmentation or additional knowledge. However, there are several deficiencies: 1) the difficulty of annotating sufficient data and introducing appropriate knowledge is considerable; 2) noisy data drive models to deviate from the target feature distribution, and unreasonable knowledge injection exacerbates the collapse of feature space of PLMs.

Recently, some prompt tuning (PT) methods \cite{Hambardzumyan2021, Li2021, Lester2021, Liu2021, Zhang2022} have shown remarkable results in low resource scenarios (i.e., parameter-efficient prompt tuning, PEPT). They freeze most or all parameters of PLMs and leverage a few additional parameters to restrict the approximation in a small manifold, thus reducing the dependency on the scale of data.

Inspired by the above, we leverage the PEPT to drive the input to match the pre-trained feature space and further present a \textbf{P}arameter-\textbf{E}fficient \textbf{M}ulti-level \textbf{I}DRR framework (\textbf{PEMI}), which alleviates the under-training problem caused by data scarcity and infuses hierarchical guidance into the prompt verbalizer. Thus we can mine better context patterns guided by hierarchical label signals for the IDRR. Generally, prompt-based framework mostly consists of two parts: \textbf{template engineering} and \textbf{verbalizer engineering}.

For the template formulation, instead of manually designed templates, we inject soft prompts into the template and regard them as learnable global context vectors to mine the unique pattern of arguments and adjust input features to align the target distribution under the pre-trained semantic space. 

However, this alignment is marginal, so it is crucial to adopt the verbalizer for the masked language model (MLM), which maps several label words in vocab to a specific category. But these verbalizer does not have access to learn the hierarchical connection of discourse relations. Besides, existing methods \cite{Wu2020, Chen2021, Wu2022a, Wang2022} require feature alignment or extra structure (e.g., GCN, CRF), which conflicts with the hypothesis of PEPT. Therefore, we propose a novel method called hierarchical label refining (HLR) to incorporate hierarchical information into the verbalizer. In our method, only the bottom-level label words are parameterized. Others are refined from the bottom up according to the hierarchical division. And the dispersed label semantics are continuously aggregated to more generalized ones in each iteration, thus realizing the dynamic updating of the verbalizer.

Finally, our framework carries out joint learning at all levels, thus combining the intra-level label discrimination process and the inter-level hierarchical information integration process.

Our contributions are summarized as follows:
\begin{itemize}
    \item Initially leverage PEPT to drive arguments to match the pre-trained feature space and alleviate the data scarcity of IDRR from the parameter side.
    \item Propose a parameter-efficient multi-level IDRR framework, deeply infusing hierarchical label guidance into prompt tuning and jointly mining the unique patterns of arguments and labels for MIDRR.
    \item Results and visualization demonstrate the effectiveness of our framework with only 100K trainable parameters. 
\end{itemize}

\section{Related Work}
\subsection{Implicit discourse relation recognition}
We introduce deep learning methods for the IDRR \cite{Pitler2009} through two routes.

One route is \textbf{argument pair enhancement}. The early work \citep{Zhang2015, Chen2016, Qin2016, Bai2018} tends to build a heterogeneous neural network to acquire structured argument representations. Besides, other methods \citep{Liu2016a, Lan2017, Guo2018, Ruan2020, Liu2020} focus on capturing interactions between arguments. Moreover, several methods \citep{Dai2018, Kishimoto2018, Guo2020, Kishimoto2020, Zhang2021} aim at obtaining robust representations based on data augmentation or knowledge projection. However, these methods lack the exploration of relation patterns.

Another route is \textbf{discourse relation enhancement}. These methods are not only concerned with argument pairs but also discourse relations. \citet{He2020} utilizes a triplet loss to establish spatial relationships between arguments and relation representation. \citet{Jiang2021} tends to predict a response related to the target relation. Most studies \cite{Nguyen2019, Wu2020, Wu2022a} import different levels of relations to complete task understanding. However, they lack consideration of data scarcity and weaken the effectiveness of PLMs. We combine prompt tuning with hierarchical label refining to mine argument and label patterns from a multi-level perspective and adopt a parameter-efficient design to alleviate the above problems. 
\subsection{Prompt Tuning}
The essence of prompt-based learning is to bridge the gap between the MLM and downstream tasks by reformulating specific tasks as cloze questions. At present, there are some papers \citep{Xiang2022, Zhou2022} that make hand-crafted prompts to achieve promotion for IDRR. However, they require numerous experiments to obtain reliable templates.

Recently, prompt tuning (PT) \cite{Liu2022, Ding2022} is proposed to search for prompt tokens in a soft embedding space. Depending on resource scenarios, it can be mainly divided into two kinds of studies: \textbf{full prompt tuning} (FPT) and \textbf{parameter-efficient ones} (PEPT). 

With sufficient data, FPT \cite{Han2021, Liu2021a, Wu2022a} combines the parameters of PLM with soft prompts to accomplish the bidirectional alignment of semantic feature space and inputs. Among them, P-Tuning \cite{Liu2021a} replaces the discrete prompts with soft ones and adopts MLM for downstream tasks. PTR \cite{Han2021} concatenates multiple sub-templates and selects unique label word sets for different sub-prompts. 

However, in the low-resource scenario, this strategy cannot accurately depict the high-dimensional task-specific space. Therefore, PEPT methods \cite{Hambardzumyan2021, Lester2021, Li2021, Liu2021, Zhang2022, Gu2022} consider fixing the parameters of PLMs, and leverage soft prompts to map the task-specific input into unified pre-trained semantic space. For example, WARP \cite{Hambardzumyan2021} uses adversarial reprogramming to tune input prompts and the self-learning Verbalizer to achieve superior performance on NLU tasks. Prefix-Tuning \cite{Li2021} tunes PLMs by updating the pre-pended parameters in each transformer layer for NLG. In this paper, we combine PEPT with our proposed hierarchical label refining method, which not only takes full advantage of PEPT for IDRR, but also effectively integrates the extraction of hierarchical guidance into the process of prompt tuning. 

\section{Overall Framework}\label{sec:model}
\begin{figure*}
    \centering
    \includegraphics[scale=0.446]{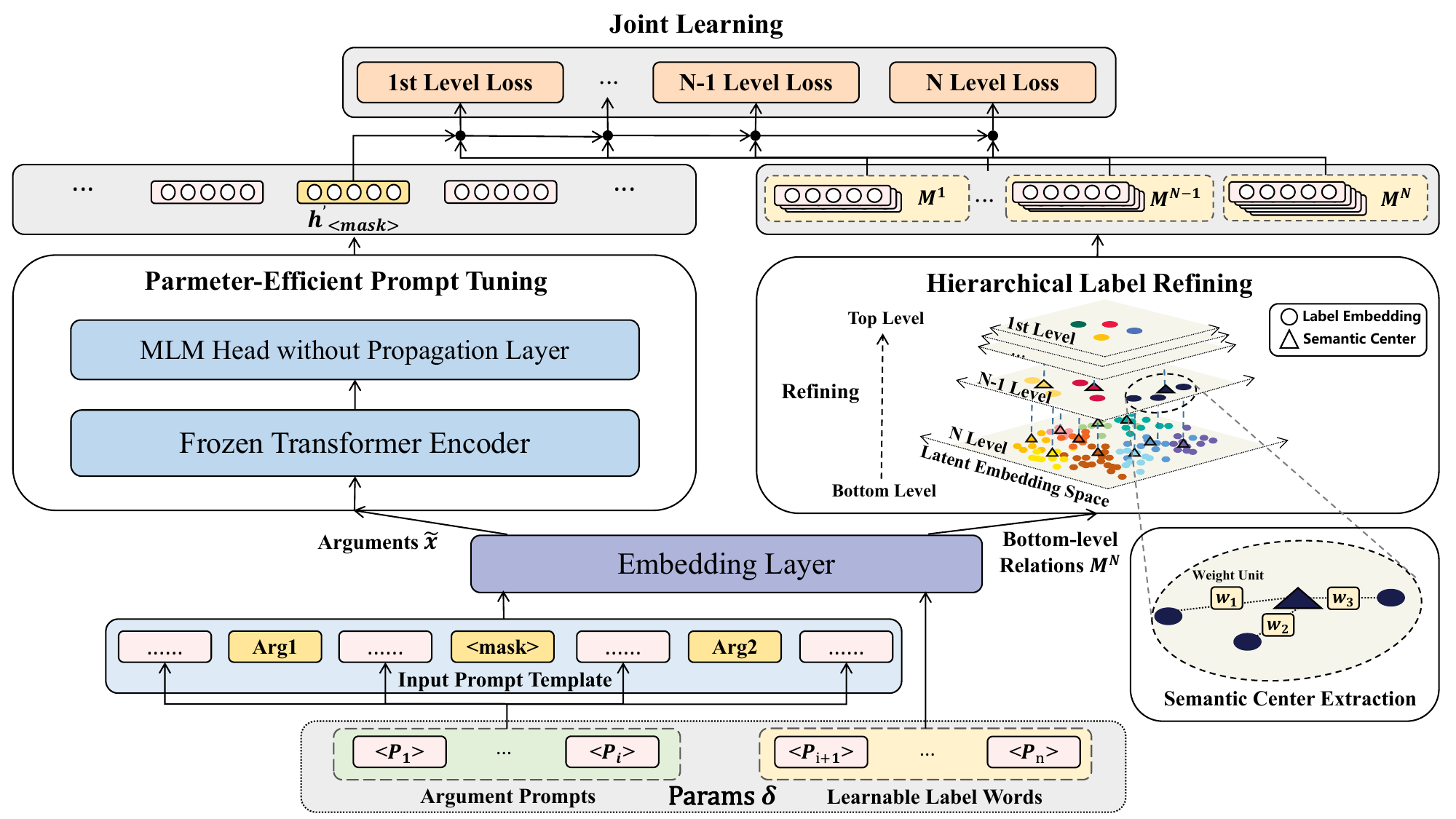}
    \caption{The overall architecture of our PEMI framework.}
    \label{fig:overall}
\end{figure*}

Let $\bm{x}=(\bm{x}_1, \bm{x}_2)\in \mathcal{X}$ be an argument pair and $\mathbb{L}=\left(L^{1},L^{2},..,L^{\mathcal{Z}}\right)$ be the set of total labels, where $L^z$ is the level-$z$ label set. The goal of the MIDRR is to predict the relation sequence $\bm{l}=l^{1},\ldots, l^{z},\ldots,l^{\mathcal{Z}}$, where $l^{z}\in L^{z}$ is the prediction of level $z$. The overview of our framework is shown in Figure \ref{fig:overall}. In this section, we explain our framework in three parts. First, we analyze the theory of PEPT for single-level IDRR and infer the association with our idea. Next, we describe how to expand the PEPT to MIDRR through our proposed hierarchical label refining method. Finally, we conduct joint learning with multiple levels so as to fuse the label information of inter and intra-levels. 
\subsection{Prompt Tuning for Single-level IDRR}\label{sec:PEPT}
Prompt tuning is a universal approach to stimulate the potential of PLMs for most downstream tasks, which goal is to find the best prompts that make the MLM predict the desired answer for the \texttt{<mask>} in templates. It is also suitable for single-level IDRR. Inspired by a PEPT method called WARP \cite{Hambardzumyan2021}, we desire to achieve objective approximations with fewer parameters for the data scarcity of IDRR. And to our knowledge, our work is the first successful application of PEPT to the IDRR.

In theory, given a MLM $\mathcal{M}$ and its vocabulary $\mathcal{V}$, it is requisite to transform $z$-th level IDRR into a MLM task. Therefore, for the input $\bm{x}$, we first construct a modified input $\tilde{\bm{x}}\in \tilde{\mathcal{X}}$ through template projection $\mathcal{T}: \mathcal{X}\rightarrow\tilde{\mathcal{X}}$, which surrounds by soft prompts $\mathcal{P}=\{\langle \mathbf{P}_0\rangle,\langle \mathbf{P}_1\rangle,...,\langle \mathbf{P}_{K-1}\rangle\}\subset \mathcal{V}$ ($K$ represents the number of $\mathcal{P}$) and special tokens \texttt{<mask>} and \texttt{<sep>}. These soft prompt tokens are the same as other words in $\mathcal{V}$. But they do not refer to any real word and are learnable through gradient backpropagation. So the actual input $\tilde{\bm{x}}\in \tilde{\mathcal{X}}$ can be formulated as follows:
\begin{equation}\label{equ:template_func}
\begin{split}
    \tilde{\bm{x}}=&\mathcal{T}(x)\\=&[\langle \mathbf{P}_{0:k_1}\rangle ,\bm{x}_1, \langle \mathbf{P}_{k_1+1:k_2}\rangle,\langle mask\rangle, \langle sep\rangle\\
    &\langle \mathbf{P}_{k_2+1:k_3}\rangle , \bm{x}_2,\langle\mathbf{P}_{k_3+1:K-1}\rangle]
\end{split}
\end{equation}
where $\langle \mathbf{P}_{0:k_1}\rangle$ indicates the aggregation of $\langle\mathbf{P}_{i}\rangle \in \mathcal{V}$ and $i\in [0, k_1]$. The value of $k_1,k_2,k_3$ is optional, and we will discuss the main factors of template selection in \ref{sec:tem_select}.\footnote{We also conduct some experiments on the position of the \texttt{<mask>} in Appendix \ref{sec:appendixB}, and our results show that it is better to place it in the middle of the two arguments.}

Then, we hope to leverage the MLM $\mathcal{M}$ to predict discourse relations. We denote $\mathcal{E}:\tilde{\mathcal{X}}\rightarrow \mathcal{H}$ and $\mathcal{F}:\mathcal{H}\rightarrow \mathcal{V}$ as the encoder and vocabulary classifier of $\mathcal{M}$. For encoder $\mathcal{E}$, we do not make extra modifications and obtain the feature representation $\mathbf{h}_{\langle mask\rangle}\in \mathcal{H}$ from \texttt{<mask>} position. Through the attention in $\mathcal{E}$, prompts can constantly mine the context pattern and guide to acquire semantic representations with IDRR characteristics. While for $\mathcal{F}$, label word selection should be made to constrain the probabilities to fall on words associated with relation labels. Here, instead of picking up verbalizer through handcraft or rules, we adopt self-learning verbalizer $\mathcal{V}^z=\{\langle\bm{V}_0\rangle,\langle\bm{V}_1\rangle,...,\langle\bm{V}_{|L^{z}|}\rangle\}\subset \mathcal{V}$ to represent label words for level-$z$ classes. We denote this new projection as $\mathcal{F}^z:\mathcal{H}\rightarrow \mathcal{V}^z$. In practice, we replace the final projection in $\mathcal{F}$ with verbalizer embedding matrix $\mathbf{M}^z\in \mathbb{R}^{|L^{z}|\times d}$ to acquire $\mathcal{F}^z$. The matrix $\mathbf{M}^z$ represents as:
\begin{equation}
    \mathbf{M}^z=\left[\bm{e}(\langle\bm{V}_0\rangle);\bm{e}(\langle\bm{V}_1\rangle);\ldots;\bm{e}(\langle\bm{V}_{|L^{z}|}\rangle)\right]
\end{equation}
where $\bm{e}(\cdot)$ is the embedding projection of $\mathcal{M}$. And the calculation of $\mathcal{F}^z$ is as follows:
\begin{equation}
    \mathcal{F}^z(\bm{h}_{\langle mask\rangle})=\hat{\mathbf{y}}^z=\textrm{softmax}(\mathbf{M}^z \mathbf{h}^{'}_{\langle mask\rangle})\label{equ:mlm_head}
\end{equation}
where $\hat{\mathbf{y}}^z=\{\hat{y}_i^z\}^{|L^z|}_{i=1}$ is the probabilistic predictions of the $z$-th level and $\mathbf{h}^{'}_{\langle mask\rangle}$ is the representation before verbalizer projection. There are different operations for each PLM (e.g., layer norm).

Finally, we train this model through cross-entropy loss to approximate the real distribution of $z$-th level as follows:
\begin{equation}
    \mathcal{L}^z=-\sum_{i=1}^{|L^z|}y^z_i\log(\hat{y}^z_i)\label{equ:main_loss}
\end{equation}
where $\mathbf{y}^z=\{y_i^z\}^{|L^z|}_{i=1}$ is the one-hot representation of ground-truth relation.

Although we can narrow the gap between pre-training and IDRR by the above, it is inappropriate to fine-tune the pre-trained feature space to task-specific ones in low-resource scenarios, which will further aggravate the collapse of the vanilla space.
Therefore, we propose to approximate the original objective by adjusting the input to fit vanilla PLM space. Let $\theta_{\mathcal{M}}$ be the parameters of $\mathcal{M}$ and $\delta=\{\theta_P,\theta_{\mathcal{V}^z}\}$ represents the parameters of soft prompts and verbalizer. Our method seeks to find a new approximate objective function $\hat{\mathcal{L}}^z(\cdot;\delta)$, such that:
\begin{equation}
    |\mathcal{L}^z(x,y;\theta_\mathcal{M},\delta)-\hat{\mathcal{L}}^z(x,y;\delta)|< \epsilon
\end{equation}
where $\epsilon$ is the approximation error. 

Moreover, if we assume that the difference of $\mathcal{F}^z$ between $\mathcal{L}^z$ and $\hat{\mathcal{L}}^z$ is insignificant when $\mathcal{L}^z$ reaches its optimal, the purpose of PEPT can be understood as:
\begin{equation}
    \mathcal{E}(\mathcal{T}(x;\theta_{\mathcal{P}}))\rightarrow\mathcal{E}^{+}(\mathcal{T}(x;\theta_{\mathcal{P}});\theta_{\mathcal{M}})
\end{equation}
where $\mathcal{E}^+$ is the optimal encoder.
Through this method, we restrict MLM into a small manifold in the functional space \cite{Aghajanyan2021}, thus adjusting the input to fit the original PLM feature space. Especially in low-resource situations, this approach can effectively achieve approximation. 

\subsection{Hierarchical Label Refining}\label{sec:hlr}
Despite the success of single-level IDRR, PEPT suffers from the absence of hierarchical label guidance. Besides, existed hierarchical infusion method \cite{Wu2020, Chen2021, Wu2022a, Wang2022} undoubtedly introduces additional parameters except for $\delta$, which accelerates the deconstruction of pre-trained feature space. 

Therefore, we propose a hierarchical label refining (HLR) method that integrates hierarchical guidance on the verbalizer. Not only does our method not increase the scale of $\theta_{\mathcal{V}}=\{\theta_{\mathcal{V}^m}\}_{m=1}^Z$, but also restrict the parameters to $\theta_{\mathcal{V}^Z}$.

In detail, for multi-level IDRR, the arguments are annotated by different semantic granularity in the process of labeling. And all the labels can form a graph $\mathcal{G}$ with $Z$ levels according to the pre-defined relationships. In this graph, for a particular $z$-th level label $l_{j}^{z}$ ($j\in \{1, 2,...,|L^z|\}$), its relevant sub-labels are distributed in level z+1 and we denote them as:
\begin{equation}
\begin{split}
   L^{z+1}_j=&\{l^{z+1}|Parent\left(l^{z+1}_t\right)=l_{j}^{z}\}\\
   &where\ t\in \{1,2,\ldots,|L^{z+1}|\}
\end{split}
\end{equation}
where $Parent(\cdot)$ means the parent node of it.

In abstract, the nodes in $L^{z+1}_j$ are the semantic divisions of $l_{j}^{z}$, which represent the local meaning of $l_{j}^{z}$. In other words, the meaning of $l_{j}^{z}$ can be extracted by merging its sub-labels. While in the embedding space, this relationship can be translated into clustering, where $l_{j}^{z}$ represents the semantic center of its support set $L^{z+1}_j$. Therefore, if the embeddings for sub-labels make sense, we can regard the semantic center extracted by them as their parent label. Under this concept, we only need to build the semantics of the bottom-level labels, and other levels are produced by aggregation from the bottom up. 
From the view of the graph neural networks, our method limits the adjacent nodes of each node in $\mathcal{G}$ to be the fine-grained labels of the first-order neighborhood, and the updating of node embeddings only depends on the aggregation of the adjacent nodes without itself. 
In practice, the verbalizer $\mathcal{V}^*$ only consists of $|L^\mathcal{Z}|$ learnable label words and others are generated from $\mathcal{V}^*$. 

Furthermore, we discuss how to achieve effective semantic refining. A major direction is the proportion of support nodes. However, the weights of refining depend on numerous factors, e.g., the label distribution of datasets, the semantic importance of the parent label, polysemy and so on.\footnote{We conducted several experiments followed by some methods \cite{Cui2019, Li2020b, Subramanian2021}, but they did not work well on our model.} Hence we apply several learnable weight units in the process of refining to balance the influence of multiple factors, which is equal to adding weights to the edges in $\mathcal{G}$. All the weights are acquired through the iteration of prompt tuning. Formally, the element of the weight vector $\bm{w}_{j}^{z}=\left[w_{j, i}^{z}\right]_{i=1}^{|L^{z+1}|}$ for $l_{j}^{z}$ are obtained as follows:
\begin{equation}
    w_{j, i}^{z} = \left\{\begin{array}{ll}
        unit(z, i, j) & l_i^{z+1}\in L_{j}^{z+1}\\
        0 & otherwise\\
    \end{array}\right.
\end{equation}
where $unit(*)$ is the function to obtain the target weight unit controlled by $z$, $i$, and $j$.

After that, We formalize the calculation of the verbalizer matrix $\bm{M}^z$ at $z$-th level as follows: 
\begin{equation}
\renewcommand{\arraystretch}{1.5}
    \mathbf{M}^z=\left\{\begin{array}{ll}
        \left[\bm{e}(\langle\bm{V}_0\rangle);\ldots;\bm{e}(\langle\bm{V}_{|L^{\mathcal{Z}}|}\rangle)\right] & z=\mathcal{Z} \\
         f(\mathbf{W}^z)\cdot \mathbf{M}^{z+1}& otherwise
    \end{array}\right.
\end{equation}
where $\mathbf{W}^z=\left[\mathbf{w}_{1}^{z};\bm{w}_{j}^{z};...;\bm{w}_{|L^z|}^{z}\right]$ is the weight matrix of $z$-th level, and $f(\cdot)$ stands for the normalization method like softmax and $L_1$ norm.

Our method repeats this process from the bottom up to get semantic embeddings at all levels. And it is performed in each iteration before the calculation of the objective function, thus aggregating upper semantics according to more precise basic ones and infusing it into the whole process of PT. In this way, discourse relations produce hierarchical guidance from the generation process and continue to enrich the verbalizer $\mathcal{V}^*$.

\subsection{Joint Learning}
After the embeddings of all levels are generated vertically, we conduct horizontal training for intra-level senses. Precisely, we first calculate the probability distribution of each level independently. The calculations of each level follow Equation \eqref{equ:mlm_head} and \eqref{equ:main_loss}.

Eventually, our model jointly learns the overall loss functions as the weighted sum of Equation \eqref{equ:main_loss}:
\begin{equation}\label{equ:all-loss}
    \mathcal{L}=\sum_{t=1}^{Z}\lambda_z \mathcal{L}^{(z)}
\end{equation}
where $\lambda_z$ indicates the trade-off hyper-parameters to balance the loss of different levels. By joint learning for different levels, our model naturally combines the information within and between hierarchies. Besides, it can synchronously manage all the levels through one gradient descent, without multiple iterations like the sequence generation model, thus speeding up the calculation while keeping hierarchical label guidance information. 

\section{Experiments}






\subsection{Dataset}
To facilitate comparison with previous work, we evaluate our model on PDTB 2.0 and 3.0 datasets. The original benchmark \cite{Prasad2008} contains three-level relation hierarchies. However, the third-level relations cannot conduct classification due to the lack of samples in most of the categories. Following previous work \citep{Wu2020, Wu2022a}, we regard the connectives as the third level for MIDRR. The PDTB 2.0 contains 4 (Top Level), 16 (Second Level) and 102 (Connectives) categories for each level. For the second-level labels, five of them without validation and test instances are removed. For PDTB 3.0, following \citet{Kim2020}, we conduct 4-way and 14-way classifications for the top and second levels. Since previous work has not defined the criterion for PDTB 3.0 connectives, we choose 150 connectives in implicit instances for classification\footnote{\url{https://github.com/cyclone-joker/IDRR_PDTB3_Conns}}. For data partitioning, we conduct the most popular dataset splitting strategies PDTB-Ji \cite{Ji2015}, which denotes sections 2-20 as the training set, sections 0-1 as the development set, and sections 21-22 as the test set. More details of the PDTB-Ji splitting are shown in Appendix \ref{sec:appendixA}.
\subsection{Experimental Settings}
Our work uses \texttt{Pytorch} and \texttt{Huggingface} libraries for development, and also verifies the effectiveness of our model on \texttt{MindSpore} library. For better comparison with recent models, we apply RoBERTa-base \cite{Liu2019} as our encoder. All of the hyper-parameters settings remain the same as the original settings for it, except for the dropout is set to 0. And we only updates the parameters of $\delta=\{\theta_{\mathcal{P}}, \theta_{\mathcal{V}^{\mathcal{Z}}}\}$ and weight units $\{\bm{W}^z\}^\mathcal{Z}_{z=1}$ while freezing all the other parameters when training. The weight coefficients of loss function $\lambda_z$ are 1.0 equally. And the normalized function $f$ is softmax. In order to verify the validity of the results, we choose Adam optimizer and learning rate 1$e$-3 with a batch size of 8. The training strategy conducts early stopping with a maximum of 15 epochs and chooses models based on the best result on the development set. The evaluation step is 500. In practice, one training process of PEMI takes about 1.5 hours on a single RTX 3090 GPU. Finally, We choose the macro-$F_1$ and accuracy as our validation metrics.
\subsection{The Comparison Models}\label{sec:baselines}
\begin{table*}[!t]
\newcommand{\tabincell}[2]{\begin{tabular}{@{}#1@{}}#2\end{tabular}}
\setlength\tabcolsep{0.95mm}
\centering
\begin{tabular}{ccccccccc}
\toprule
\multirow{2}{*}{\textbf{Model}} & \multirow{2}{*}{\tabincell{c}{\textbf{Embedding}\\\textbf{Layer}}} & \multicolumn{2}{c}{\tabincell{c}{\textbf{Top Level}\\\textbf{(4-way)}}} &\multicolumn{2}{c}{\tabincell{c}{\textbf{Second Level}\\\textbf{(11-way)}}}&\multicolumn{2}{c}{\tabincell{c}{\textbf{Connectives}\\\textbf{(102-way)}}}& \multirow{2}{*}{\tabincell{c}{\textbf{Trainable}\\\textbf{Params}}}  \\
    &   & \bm{$F_1$}&\bm{$Acc$} & \bm{$F_1$}&\bm{$Acc$} &\bm{$F_1$}&\bm{$Acc$}\\
\hline
FT-RoBERTa \cite{Liu2019}&RoBERTa&61.62&68.57&38.55&58.43&7.89&29.68&>125M\\
BMGF \cite{Liu2020}&RoBERTa&63.39&69.06&-&58.13&-&-&>15M\\
\hline
MTL-KT \cite{Nguyen2019}&RoBERTa&61.89&68.42&38.10&57.72& 7.75&29.57&>125M\\
MT-BERT \cite{Kishimoto2020}&BERT&58.48&65.26&-&54.32&-&-&>110M\\
TransS-RoBERTa \cite{He2020}&RoBERTa&61.57&69.28&37.83&57.76&7.83&31.38&>125M\\
HierMTN-CRF \cite{Wu2020}&BERT&55.72&65.26&33.91&52.34&10.37&30.00&>110M\\
HierMTN-CRF \cite{Wu2020}&RoBERTa&62.02&70.05&38.28&58.61&10.45&31.30&>125M\\
CG-T5 \cite{Jiang2021}&T5&57.18&-&37.76&-&-&-&>250M\\
LDSGM \cite{Wu2022a}&RoBERTa&63.73&\textbf{71.18}&40.49&60.33&10.68&32.20&>155M\\
\hline
\textbf{Ours}&RoBERTa&\textbf{64.05}&71.13&\textbf{41.31}&\textbf{60.66}&\textbf{10.87}&\textbf{35.32}&\textbf{<100K}\\
\bottomrule
\end{tabular}
\caption{Experimental results for Macro-$F_1$ score (\%), Accuracy (\%) and Trainable Parameters on PDTB 2.0. The results of FT-RoBERTa and TransS-RoBERTa are obtained under our settings.}
\label{tab:main}
\end{table*}
\begin{table}
\setlength\tabcolsep{0.8mm}
    \begin{tabular}{cccc}
    \toprule
        \multirow{2}{*}{\textbf{Second Level}} & \multicolumn{3}{c}{\textbf{Label-wise F1 (\%)}}\\
        &\textbf{BMGF} & \textbf{LDSGM} & \textbf{Ours}\\
        \hline
        \textit{Comp}.Concession &0&0&\textbf{8.11}\\
        \textit{Comp}.Contrast &59.75&\textbf{63.52}&60.20\\
        \hline
        \textit{Cont}.Cause &59.60&\textbf{64.36}&61.82\\
        \textit{Cont}.Pragmatic cause &0&0&0\\
        \hline
        \textit{Expa}.Alternative &60.00&\textbf{63.46}&60.54\\
        \textit{Expa}.Conjunction&\textbf{60.17}&57.91&50.71\\
        \textit{Expa}.Instantiation &69.96&72.6&\textbf{73.81}\\
        \textit{Expa}.List&0&8.98&\textbf{30.55}\\
        \textit{Expa}.Restatement &53.83&\textbf{58.06}&55.60\\
        \hline
        \textit{Temp}.Asynchronous &56.18&\textbf{56.47}&53.04\\
        \textit{Temp}.Synchrony &0&0&0\\
    \bottomrule
    \end{tabular}
    \caption{The second-level label-wise $F_1$ on PDTB 2.0. \textit{Comp}, \textit{Cont}, \textit{Expa} and \textit{Temp} represents Comparison, Contingency, Expansion and Temporal separately.}
    \label{tab:second_results}
\end{table}
In this section, we select some baselines for PDTB 2.0 and 3.0 separately and introduce them briefly: \\
$\bullet\ $\textbf{PDTB 2.0 :} We select some comparable models based on PLMs and briefly introduce them through two aspects: \\
\textbf{Argument Pair Enhancement}

1) \textbf{FT-RoBERTa}: \citet{Liu2019} improves the BERT by removing the NSP task and pre-training on wide corpora. We conduct experiments for each level separately. 

2) \textbf{BMGF}: \citet{Liu2020} proposes a bilateral multi-perspective matching encoder to enhance the arguments interaction on both text span and sentence level. \\
\textbf{Discourse Relation Enhancement}

3) \textbf{MTL-KT}: \citet{Nguyen2019} predicts relations and connectives simultaneously and transfers knowledge via relations and connectives through label embeddings. We import the RoBERTa version from \citet{Wu2022a}. 

4) \textbf{MT-BERT}: \citet{Kishimoto2020} proposes a multi-task learning model which additionally predicts connectives and explicit discourse relations and adds extra data. 

5) \textbf{TransS-RoBERTa}: \citet{He2020} uses triplet loss to introduce geometric structure into semantic representation space. We replace the embedding layer with RoBERTa for a fair comparison. 

6) \textbf{HierMTN-CRF}: \citet{Wu2020} firstly deals with multi-level IDRR simultaneously and chooses the label sequence based on a CRF layer. We import its BERT and RoBERTa versions. 

7) \textbf{CG-T5}: \citet{Jiang2021} combines the IDRR classification with generation by generating adequate sentences related to discourse relations with several templates. 

8) \textbf{LDSGM}: \citet{Wu2022a} views IDRR as a label sequence prediction task and leverages the label dependencies between discourse relations through GCN and conducts label sequence prediction by a GRU decoder. \\
$\bullet\ $\textbf{PDTB 3.0 :}

1) \textbf{NNMA}: \citet{Liu2016a} imitates repeat reading habit by applying stacked attention mechanisms on the representations of argument pair. 

2) \textbf{MANN}: \citet{Lan2017} regards the IDRR for multiple datasets as multi-task learning and applies interactive attention based on BiLSTM. 

3) \textbf{IPAL}: \citet{Ruan2020} divides argument pair encoding into two channels and combines self-attention and interactive attention by a cross-coupled network. 

4) \textbf{MANF}: \citet{Xiang2022a} proposes dual attention and encodes word-pairs offsets to enhance semantic interaction. We import the word2vec and BERT versions of it. 

5) \textbf{FT-RoBERTa}: we also fine-tune a RoBERTa model on PDTB 3.0 for better comparison. 
\subsection{Results and Analysis}
\begin{table}
\setlength\tabcolsep{1.5mm}
\centering
\begin{tabular}{cccc}
\toprule
\multirow{2}{*}{\textbf{Model}}& \multicolumn{3}{c}{\textbf{Macro-F1}}\\
&\textbf{Top}&\textbf{Second}&\textbf{Conn}\\
\hline
Baseline&61.29&39.19&8.12\\
+PEPT&63.16&40.71&9.89\\
+HLR&62.85&40.82&8.94\\
+PEPT\&HLR (Ours)&\textbf{64.05}&\textbf{41.31}&\textbf{10.87}\\
\bottomrule
\end{tabular}
\caption{Ablation study on PDTB 2.0. Our \textbf{Baseline} choose fine-tuned RoBERTa MLM with a learnable verbalizer. \textbf{PEPT} means parameter-efficient prompt tuning and \textbf{HLR} is the hierarchical label refining.}
\label{tab:ablation}
\end{table}
In this section, we display the main results of three levels on PDTB 2.0 (Table \ref{tab:main}) and PDTB 3.0 (Table \ref{tab:PDTB3_main}) and the label-wise $F_1$ of level 2 on PDTB 2.0 (Table \ref{tab:second_results}) and PDTB 3.0 (Table \ref{tab:PDTB3_second_results}). 

We can obtain the following observations from these results: \textbf{1)} In table \ref{tab:main}, our model achieves comparable performance with strong baselines and only uses 0.1\% trainable parameters. And the improvement mainly occurs at the level-3 senses, which states that our model is more aware of fine-grained hierarchical semantics. \textbf{2)} In table \ref{tab:PDTB3_main}, compared with baselines, our model exceeds all fine-tuned models currently, which proves that the effect of our model is also guaranteed with sufficient data. \textbf{3)} In Table \ref{tab:second_results}, our model mainly outperforms on the minor classes. For PDTB 2.0, the improvement depends on three minor categories: \emph{Comp.Concession}, \emph{Expa.List} and \emph{Expa.Instantiation},  which indicates that the approximation through fewer trainable parameters drives the model to pay more attention to minors. More details for PDTB 3.0 are shown in Appendix \ref{sec:pdtb3}.
\subsection{Ablation Study and Analysis}
We conduct the ablation study on PDTB 2.0 to deeply analyze the impact of our framework. Our \textbf{Baseline} chooses fine-tuned RoBERTa MLM with a learnable verbalizer. Compared with fine-tuned RoBERTa, our baseline acquires arguments representation through \texttt{<mask>} and retains some parameters of MLM head. Besides, it treats IDRR of different levels as an individual classification but shares the parameters of the encoder. And then, we decompose our model into two parts described in Section \ref{sec:model}: Parameter-Efficient Prompt Tuning (\textbf{PEPT}) and hierarchical label refining (\textbf{HLR}). 

From Table \ref{tab:ablation}, we can observe that: \textbf{1)} The results of our baseline are higher than the vanilla PLM, which indicates that adapting MLM to the IDRR is more practicable. \textbf{2)} Baseline+HLR gains improvements on all levels, especially on level 2, which presumes that information from both the upper and lower level labels guides to make it more semantically authentic. \textbf{3)} PEMI achieves the best performance over other combinations, which proves that  PEPT makes HLR not be affected by redundant parameters and focuses on the semantic information in the verbalizer.

\subsection{Template Selection}\label{sec:tem_select}
\begin{figure}
    \centering
    \includegraphics[scale=0.47]{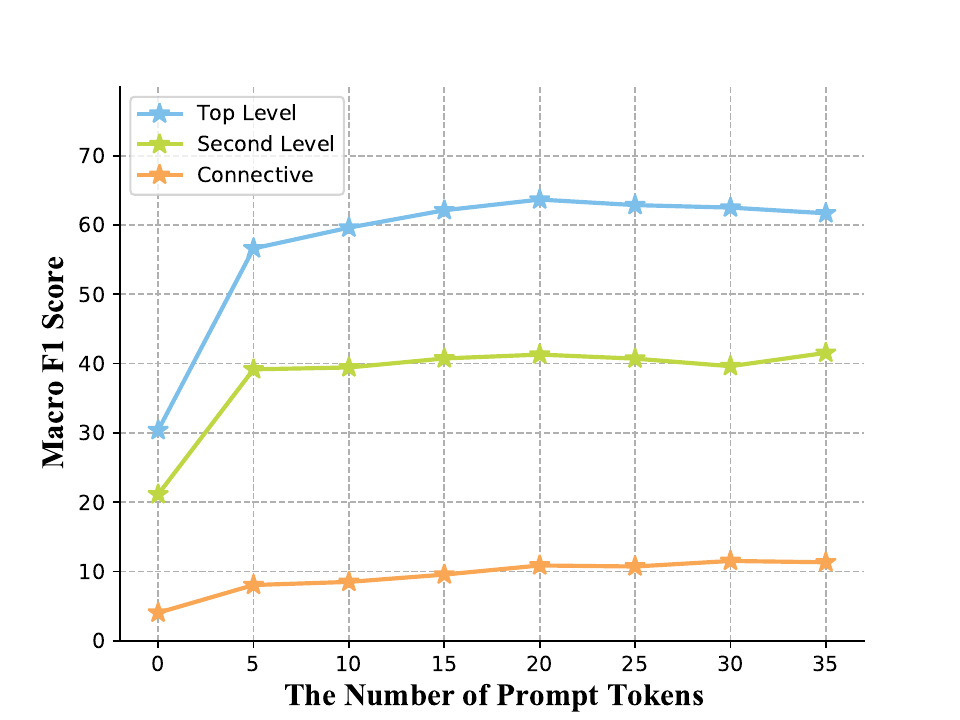}
    \caption{The effect of prompt token size for MIDRR on PDTB 2.0. We follow the best template in Table \ref{tab:all_input_location} and try to put them uniformly in each location.}
    \label{fig:prompt_size}
\end{figure}
Furthermore, we design experiments on PDTB 2.0 for two main factors of the prompt templates: the \textbf{location} and the \textbf{size} of prompt tokens, as shown in Table \ref{tab:all_input_location} and Figure \ref{fig:prompt_size} separately. 

Table \ref{tab:all_input_location} shows that the locations have a great influence on our model. Generally, we note that most of the templates that prompt tokens are scattered surpass the compact ones. So it is beneficial to place scattered around sentences. Meticulous, placing more prompt tokens around the first sentence achieves particular promotion, suggesting that early intervention for prompts could better guide the predictions of discourse relations.

In Figure \ref{fig:prompt_size}, as the number of prompt tokens increases, the situations are different for three levels. For the level-1 and level-2 senses, they reach the peak when the number rises to 20 and then starts to go down, which indicates that over many prompt tokens may dilute the attention between arguments. However, the performance of connectives continues to improve as the number increases. This is mainly because the difficulty of classification rises and more prompts need to be involved. Therefore, we ultimately measured the performance of all levels and chose 20 prompt tokens as our final result, but there is still room for improvement.
\subsection{Impact of Hierarchical Label Refining}
\begin{figure}
    \centering
    \includegraphics[scale=0.47]{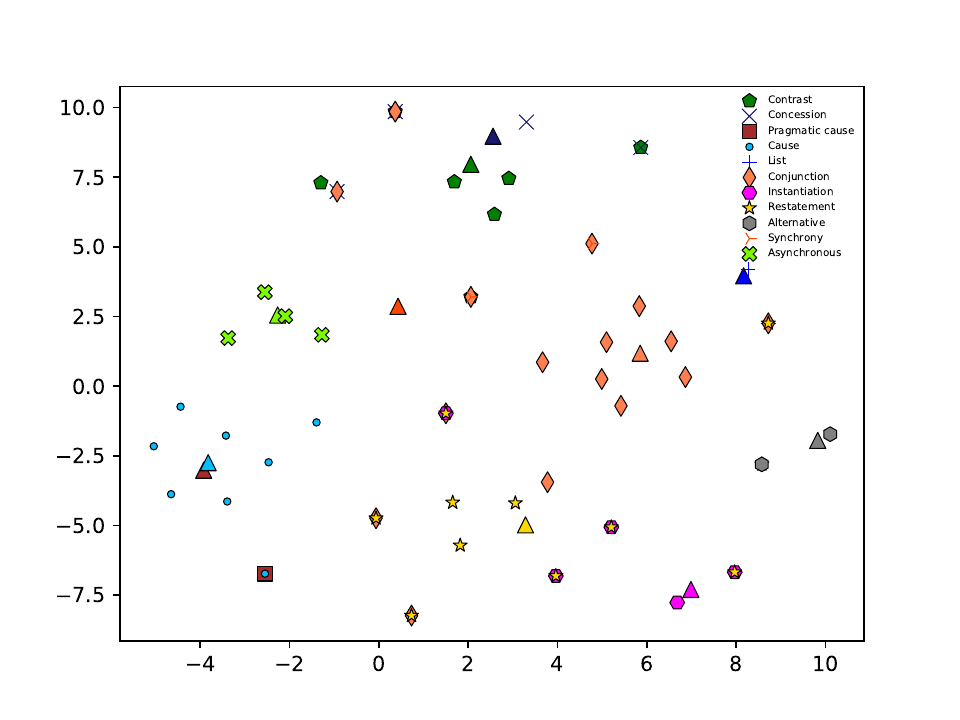}
    \caption{Visualization of HLR method for connectives. $\Delta$ represents level 2 labels and different colors indicate different classes. We use different markers since some connectives are overlapping due to the many-to-many mapping between level 2 and connectives. }
    \label{fig:refine_clustering}
\end{figure}
Finally, we carry out two experiments to explore the impact of our HLR method: weight coefficients learned by weight units in Table \ref{tab:weight_units} and \ref{tab:additional_weight_units} and visualization of label embeddings in Figure \ref{fig:refine_clustering}.

In Table \ref{tab:weight_units}, we find out that most of the weight coefficients are inversely proportional to data size, while a few cases like \emph{Expa.Alternative} are ignored. Combined with Table \ref{tab: pdtb2_second}, we can infer that our model pays more attention to the minor classes and lowers the weight to the good-performing classes.

Besides, in Figure \ref{fig:refine_clustering}, we note that visibly clustering relationships exist in the embedding space. Meanwhile, for the major classes like \textit{Cont.Cause} and \textit{Expa.Conjunction}, the class centers tend to be the average of connectives in the cluster. In contrast, minor classes like \textit{Expa.Alternative} and \textit{Expa.List} are biased towards a particular connective. The reason is that some connectives belonging to multiple discourse relations can transfer knowledge from other relations and improve the prediction of the current relation. Then the model will increase the weight of those connectives to get closer to the actual distribution. Therefore, it can be said that the HLR method transfers the inter and intra-level guidance information in the embedding space.

\section{Conclusion}
In this paper, we tackle the problem of data scarcity for IDRR from a parameter perspective and have presented a novel parameter-efficient multi-level IDRR framework, which leverages PEPT to adjust the input to match the pre-trained space with fewer parameters and infuse hierarchical label guidance into the verbalizer. Experiments show that our model adopts parameter-efficient methods while it is comparable with recent SOTA models. Besides, it indicates that our framework can effectively stimulate the potential of PLMs without any intervention of additional data or knowledge. In the future, we will further explore the linguistic features of labels and enhance the discrimination against connectives.
\section*{Limitations}
Although our model obtains satisfying results, it also exposes some limitations. \textbf{First}, for a fair comparison to other models, we mainly carry out relevant experiments on PDTB 2.0. Due to the lack of baselines on PDTB 3.0, further analysis and comparison cannot be conducted. \textbf{Second}, in our experiments, we can find out that the HLR method does not improve the top-level or bottom-level results effectively, indicating that with the increase of the level, the refining method is insufficient to continue to generalize the bottom-level labels and further improvement should be made according to the specific features of the IDRR task. \textbf{Third}, due to the limitation of space, this paper does not focus much on semantic weight for the refining of sub-labels. This is a very broad topic involving the rationality of the discourse relation annotation and the interpretability of the label embeddings. We will conduct a further study which may appear in our next work.
\section*{Acknowledgement}
Our Work is supported by the National Natural Science Foundation of China (No. 61976154) and the CAAI-Huawei MindSpore Open Fund. We also appreciate the suggestions from ACL anonymous reviewers. 
\bibliography{custom, IDRR}
\bibliographystyle{acl_natbib}
\clearpage
\appendix
\section*{Appendices}
\section{Details of PDTB-Ji Splitting}
\label{sec:appendixA}
In this section, we provide data statistics of level 2 for PDTB 2.0 (Table \ref{tab: pdtb2_second}) and PDTB 3.0 (Table \ref{tab: pdtb3_second}) separately. 
\begin{table}[!h]
\setlength\tabcolsep{2.5mm}
    \centering
    \begin{tabular}{cccc}
        \toprule
        \multirow{2}{*}{\textbf{Second Level}}&\multicolumn{3}{c}{\textbf{Sample Size}}\\
         &\textbf{Train}&\textbf{Dev}&\textbf{Test}\\
         \hline
         \textit{Comp}.Concession&183&15&17\\
         \textit{Comp}.Contrast&1607&166&128\\
         \hline
         \textit{Cont}.Cause&3270&281&269\\
         \textit{Cont}.Pragmatic cause&64&6&7\\
         \hline
         \textit{Expa}.Alternative&147&10&9\\
         \textit{Expa}.Conjuction&2872&258&200\\
         \textit{Expa}.Instantiation&1063&106&118\\
         \textit{Expa}.List&338&9&12\\
         \textit{Expa}.Restatement&2404&260&211\\
         \hline
         \textit{Temp}.Asynchronous&532&46&54\\
         \textit{Temp}.Synchronous&203&8&14\\
         \hline
         Total&12683&1165&1039\\
    \bottomrule
    \end{tabular}
    \caption{Statistics for relation senses of Level 2 in PDTB 2.0 by PDTB-Ji splitting.}
    \label{tab: pdtb2_second}
\end{table}
\begin{table}[!h]
\setlength\tabcolsep{2.5mm}
    \centering
    \begin{tabular}{cccc}
        \toprule
        \multirow{2}{*}{\textbf{Second Level}}&\multicolumn{3}{c}{\textbf{Sample Size}}\\
         &\textbf{Train}&\textbf{Dev}&\textbf{Test}\\
         \hline
         \textit{Comp}.Concession&1164&103&97\\
         \textit{Comp}.Contrast&741&82&54\\
         \hline
         \textit{Cont}.Cause&4475&448&404\\
         \textit{Cont}.Cause+Belief&159&13&15\\
         \textit{Cont}.Condition&150&18&15\\
         \textit{Cont}.Purpose&1092&96&89\\
         \hline
         \textit{Expa}.Conjuction&3586&298&236\\
         \textit{Expa}.Equivalence&254&25&30\\
         \textit{Expa}.Instantiation&1166&116&124\\
         \textit{Expa}.Level-of-detail&2601&261&208\\
         \textit{Expa}.Manner&615&14&17\\
         \textit{Expa}.Substitution&343&27&26\\
         \hline
         \textit{Temp}.Asynchronous&1007&101&105\\
         \textit{Temp}.Synchronous&435&33&43\\
         \hline
         Total&17788&1635&1463\\
         \bottomrule
    \end{tabular}
    \caption{Statistics for relation senses of Level 2 in PDTB 3.0 by PDTB-Ji splitting.}
    \label{tab: pdtb3_second}
\end{table}
\section{Experimental Results on PDTB 3.0}\label{sec:pdtb3}
Due to the limitation of pages, we provide results of PDTB 3.0 in this section. Table \ref{tab:PDTB3_second_results} displays the label-wise F1 for level-2 senses on PDTB 3.0 and Table \ref{tab:PDTB3_main} shows the main results on PDTB 3.0 compared with the baselines we stated in Section \ref{sec:baselines}.
\begin{table}[htp]
    \centering
    \begin{tabular}{cc}
        \toprule
        \multirow{2}{*}{\textbf{Second Level}}&\textbf{Label-wise F1(\%)}\\
        &\textbf{PEMI (Ours)}\\
        \hline
         \textit{Comp}.Concession&64.68  \\
         \textit{Comp}.Contrast&52.94\\
         \hline
         \textit{Cont}.Cause&69.04\\
         \textit{Cont}.Cause+Belief&0.00\\
         \textit{Cont}.Condition&68.97\\
         \textit{Cont}.Purpose&91.49\\
         \hline
         \textit{Expa}.Conjuction&58.82\\
         \textit{Expa}.Equivalence&0.00\\
         \textit{Expa}.Instantiation&70.42\\
         \textit{Expa}.Level-of-detail&54.25\\
         \textit{Expa}.Manner&59.26\\
         \textit{Expa}.Substitution&48.98\\
         \hline
         \textit{Temp}.Asynchronous&66.67\\
         \textit{Temp}.Synchronous&32.73\\
         \bottomrule
    \end{tabular}
    \caption{The second-level label-wise F1 on PDTB 3.0.}
    \label{tab:PDTB3_second_results}
\end{table}

\begin{table*}[]
\newcommand{\tabincell}[2]{\begin{tabular}{@{}#1@{}}#2\end{tabular}}
\setlength\tabcolsep{1.1mm}
    \centering
\begin{tabular}{ccccccccc}
\toprule
\multirow{2}{*}{\textbf{Model}} & \multirow{2}{*}{\tabincell{c}{\textbf{Embedding}\\\textbf{Layer}}} & \multicolumn{2}{c}{\tabincell{c}{\textbf{Top-level}\\\textbf{(4-way)}}} &\multicolumn{2}{c}{\tabincell{c}{\textbf{Second-level}\\\textbf{(14-way)}}}&\multicolumn{2}{c}{\tabincell{c}{\textbf{Connective}\\\textbf{(150-way)}}}& \multirow{2}{*}{\tabincell{c}{\textbf{Trainable}\\\textbf{Params}}}  \\
    &   & \bm{$F_1$}&\bm{$Acc$} & \bm{$F_1$}&\bm{$Acc$} &\bm{$F_1$}&\bm{$Acc$}\\
\hline
\textbf{NNMA} \cite{Liu2016a}&GloVe&46.13&57.67&-&-&-&-&>5M\\
\textbf{MANN} \cite{Lan2017}&word2vec&47.29&57.06&-&-&-&-&>1M\\
\textbf{IPAL} \cite{Ruan2020}&BERT&49.45&58.01&-&-&-&-&>110M\\
\textbf{MANF} \cite{Xiang2022a}&word2vec&53.14&60.45&-&-&-&-&>10M\\
\textbf{MANF} \cite{Xiang2022a}&BERT&56.63&64.04&-&-&-&-&>110M\\
\textbf{FT-RoBERTa} \cite{Liu2019}&RoBERTa&66.94&71.91&51.78&61.24&10.07&\textbf{40.26}&>125M\\
\hline
\textbf{Ours}&RoBERTa&\textbf{69.06}&\textbf{73.27}&\textbf{52.73}&\textbf{63.09}&\textbf{10.52}&39.92&\textbf{<130K}\\
\bottomrule
\end{tabular}
\caption{Experimental results for Macro-$F_1$ score (\%), Accuracy (\%) and Trainable Parameters on PDTB 3.0. The results of FT-RoBERTa are conducted based on our experimental settings.}
    \label{tab:PDTB3_main}
\end{table*}
\section{Selection of Input Templates}
\label{sec:appendixB}
In this section, we provide several templates by changing the location of prompt tokens and $\langle mask\rangle$ to explore the validity of IDRR. And Table \ref{tab:all_input_location} shows the overall results for reference. Finally, we find out that it is preferable to put the $\langle mask\rangle$ token in the middle of the argument pair, as described in Section \ref{sec:PEPT}.
\begin{table*}
    \centering
    \begin{tabular}{ccccccc}
    \toprule
        \multirow{2}{*}{\textbf{Template Form}} & \multicolumn{2}{c}{\textbf{Top-level}} & \multicolumn{2}{c}{\textbf{Second-level}} & \multicolumn{2}{c}{\textbf{Connective}}\\
        & \bm{$F_1$} & \bm{$Acc$} & \bm{$F_1$} & \bm{$Acc$}& \bm{$F_1$} & \bm{$Acc$}\\
        \hline
        $\langle \textrm{P:4}\rangle \bm{S_1}\langle \textrm{P:4}\rangle\langle \textrm{mask}\rangle\langle \textrm{P:4}\rangle\langle \textrm{sep}\rangle\langle \textrm{P:4}\rangle \bm{S_2}\langle \textrm{P:4}\rangle$&\textbf{64.05}&71.13&\textbf{41.31}&\textbf{60.66}&\textbf{10.87}&35.32\\
        \hline
        $\langle \textrm{P:5}\rangle\bm{S_1}\langle \textrm{P:5}\rangle\langle \textrm{mask}\rangle\langle \textrm{sep}\rangle\langle \textrm{P:5}\rangle\bm{S_2}\langle \textrm{P:5}\rangle$&62.73&68.96&41.10&58.98&10.52&34.69\\
        $\langle \textrm{P:5}\rangle\langle \textrm{mask}\rangle\bm{S_1}\langle \textrm{P:5}\rangle\langle \textrm{sep}\rangle\langle \textrm{P:5}\rangle\bm{S_2}\langle \textrm{P:5}\rangle$&59.71&67.21&37.48&55.62&8.98&33.08\\
        $\langle \textrm{P:5}\rangle \bm{S_1}\langle \textrm{P:5}\rangle\langle \textrm{sep}\rangle\langle \textrm{P:5}\rangle\bm{S_2}\langle\textrm{mask}\rangle\langle \textrm{P:5}\rangle$&60.54&68.33&37.37&56.72&9.07&34.15\\
        \hline
        $\langle \textrm{P:20}\rangle\bm{S_1}\langle \textrm{mask}\rangle\langle \textrm{sep}\rangle\bm{S_2}$&63.62&\textbf{71.68}&38.59&59.44&10.57&35.37\\
        $\langle \textrm{P:20}\rangle\bm{S_1}\langle \textrm{sep}\rangle\bm{S_2}\langle \textrm{mask}\rangle$&58.66&67.95&37.73&56.67&8.61&33.33\\
        $\langle \textrm{P:20}\rangle\langle \textrm{mask}\rangle\bm{S_1}\langle \textrm{sep}\rangle\bm{S_2}$&59.32&68.76&38.59&57.91&7.91&32.28\\
        \hline
        $\bm{S_1}\langle \textrm{mask}\rangle\langle \textrm{sep}\rangle\bm{S_2}\langle \textrm{P:20}\rangle$&61.91&69.32&40.30&57.80&9.88&35.12\\
         $\langle \textrm{mask}\rangle\bm{S_1}\langle \textrm{sep}\rangle\bm{S_2}\langle \textrm{P:20}\rangle$&50.38&62.79&35.20&51.80&5.52&27.67\\
         $\bm{S_1}\langle \textrm{sep}\rangle\bm{S_2}\langle \textrm{mask}\rangle\langle \textrm{P:20}\rangle$&55.46&63.98&37.99&53.54&6.58&28.45\\
         \hline
        $\langle \textrm{P:10}\rangle \bm{S_1}\langle \textrm{mask}\rangle\langle \textrm{sep}\rangle \bm{S_2}\langle \textrm{P:10}\rangle$&62.37&69.09&39.47&57.00&9.31&34.87\\
        $\bm{S_1}\langle \textrm{P:10}\rangle\langle\textrm{mask}\rangle\langle \textrm{sep}\rangle\langle\textrm{P:10}\rangle\bm{S_2}$&59.43&67.89&38.47&56.00&8.14&34.23\\
        $\langle \textrm{P:10}\rangle\langle \textrm{mask}\rangle \bm{S_1}\langle \textrm{sep}\rangle \bm{S_2}\langle \textrm{P:10}\rangle$&59.60&68.11&37.57&58.22&8.55&32.73\\
        $\langle \textrm{P:10}\rangle \bm{S_1}\langle \textrm{sep}\rangle \bm{S_2}\langle \textrm{mask}\rangle\langle \textrm{P:10}\rangle$&60.23&68.07&37.88&58.42&8.71&32.69\\
        \hline
        $\langle \textrm{P:5}\rangle\bm{S_1}\langle \textrm{mask}\rangle\langle \textrm{sep}\rangle\bm{S_2}\langle\textrm{P:15}\rangle$&63.36&69.10&36.81&58.71&9.04&\textbf{37.31}\\
        $\langle \textrm{P:5}\rangle\langle \textrm{mask}\rangle\bm{S_1}\langle \textrm{sep}\rangle\bm{S_2}\langle\textrm{P:15}\rangle$&60.32&68.21&37.50&57.95&9.07&35.11\\ 
        $\langle \textrm{P:5}\rangle\bm{S_1}\langle \textrm{sep}\rangle \bm{S_2}\langle \textrm{mask}\rangle\langle\textrm{P:15}\rangle$&60.89&67.08&31.61&51.97&8.62&28.73\\
        \hline
        $\langle \textrm{P:15}\rangle\bm{S_1}\langle \textrm{mask}\rangle\langle \textrm{sep}\rangle\bm{S_2}\langle\textrm{P:5}\rangle$&63.03&69.86&39.88&59.74&10.73&35.89\\
         $\langle \textrm{P:15}\rangle\langle \textrm{mask}\rangle\bm{S_1}\langle \textrm{sep}\rangle \bm{S_2}\langle\textrm{P:5}\rangle$&60.77&68.51&38.07&58.41&9.57&37.11\\ 
         $\langle \textrm{P:15}\rangle\bm{S_1}\langle \textrm{sep}\rangle \bm{S_2}\langle \textrm{mask}\rangle\langle\textrm{P:5}\rangle$&61.72&69.55&38.93&59.28&9.49&33.09\\
    \bottomrule
    \end{tabular}
    \caption{Results by changing the locations of prompt tokens and $\langle mask\rangle$ on PDTB 2.0. We fix the size of the prompt tokens as 20 and test some of extreme cases based on simple permutations. $\langle \textrm{P:x}\rangle$ represents that there are $\textrm{x}$ prompt tokens inserted on this location.}
    \label{tab:all_input_location}
\end{table*}
\section{Details of Weignt Units}\label{sec:appendixC}
In this section, we display weight coefficients learned by weight units in section \ref{sec:hlr}, as shown in Table \ref{tab:weight_units} and \ref{tab:additional_weight_units}. We can observe some characteristics of the weights learned by the units. Comparing Table \ref{tab: pdtb2_second} and \ref{tab:weight_units}, it is apparent that the weight is inversely proportional to the number of samples, which suggests that our model intentionally learns features from minor classes. While for the second level, the situation is complicated. Some minor connectives like "meanwhile" in \emph{Expa.List} are put high weight and others like "furthermore" are quite the opposite. Therefore, is not enough to learn a good weight from sample size. Besides, since connectives can belong to different labels, the semantics learned from other relations can be beneficial for the current ones. 
\begin{table}[H]
\setlength\tabcolsep{1.0mm}
\newcommand{\tabincell}[2]{\begin{tabular}{@{}#1@{}}#2\end{tabular}}
    \centering
    \begin{tabular}{c|c}
    \toprule
    \textbf{Label}&\textbf{Sub Label} (\textbf{Weight} (\%))\\
    \hline
    \textit{Comp}&Contrast (51.83), Concession (48.17)\\
    \hline
    \textit{Cont}&Pragmatic cause (70.35), Cause (29.65)\\
    \hline
    \textit{Expa}&\tabincell{c}{Alternative (0.66), Conjunction (46.07),\\\ Instantiation (6.67), List (45.60),\\\ Restatement (1.01)}\\
    \hline
    \textit{Temp}&Synchrony (60.16), Asynchronous (39.84)\\
    \bottomrule
    \end{tabular}
    \caption{Weights between top and second levels.}
    \label{tab:weight_units}
\end{table}

\begin{table}[H]
\setlength\tabcolsep{0.9mm}
\newcommand{\tabincell}[2]{\begin{tabular}{@{}#1@{}}#2\end{tabular}}
    \centering
    \begin{tabular}{c|c}
    \toprule
    \textbf{Label}&\textbf{Sub Label (Weight (\%))}\\
    \hline
    \textbf{Concession}&\tabincell{c}{while(4.91), however(3.55),\\\ 
    but(3.66), even though(12.28),\\\ 
    nevertheless(7.17), still(5.41),\\\ 
    nonetheless(31.82), yet(4.65),\\\ 
    in fact(4.60), although(3.61),\\\ 
    by comparison(18.35)}\\
    \hline
    \textbf{\tabincell{c}{Pragmatic\\\ cause}}&\tabincell{c}{because(0.77), as(0.84),\\\
    in fact(1.24), since(2.10),\\\
    inasmuch as(86.28), so(1.99)\\\
    for example(1.06), thus(2.19),\\\
    for instance(1.09), indeed(2.45)}\\
    \hline
    \textbf{List}&\tabincell{c}{and(11.28), first(9.24),\\\
    while(4.89), second(2.99),\\\
    finally(13.82), in addition(4.83),\\\
    also(3.39), meanwhile(17.53),\\\
    third(2.54), furthermore(2.80),\\\
    for instance(3.09), in fact(5.05),\\\
    although(18.56)}\\
    \hline
    \textbf{Instantiation}&\tabincell{c}{indeed(4.37), for instance(9.94),\\\
    first(4.29), specifically(4.78),\\\
    in fact(4.63), for example(6.64),\\\
    for one thing(16.44), and(5.75),\\\
    for one(3.01), in particular(3.85),\\\
    on one hand(18.69), as(17.61)}\\
    \hline
    \textbf{Synchrony}&\tabincell{c}{meanwhile(6.61), while(6.40),\\\ 
    at the time(7.43), when(9.18),\\\
    as(5.49), at that time(3.86),\\\
    then(4.13), and(13.22),\\\
    simultaneously(16.46),\\\
    in the meantime(13.07),\\\
    at the same time(14.14)}\\
    \bottomrule
    \end{tabular}
    \caption{Partial weights between second-level and connectives.}
    \label{tab:additional_weight_units}
\end{table}
\end{document}